\documentclass[letterpaper,journal]{IEEEtran}
\usepackage{amsmath,amsfonts}
\usepackage{algorithmic}
\usepackage{algorithm}
\usepackage{array}
\usepackage[caption=false,font=normalsize,labelfont=sf,textfont=sf]{subfig}
\usepackage{textcomp}
\usepackage{stfloats}
\usepackage{url}
\usepackage{verbatim}
\usepackage{graphicx}
\usepackage{cite}
\usepackage{cuted}
\usepackage{capt-of}
\usepackage{booktabs}
\usepackage{cite}
\usepackage[hidelinks]{hyperref}
\newcommand{\pmv}[2]{#1\,{\scriptsize$\pm$#2}}
\usepackage[font=footnotesize,labelfont=sc,labelsep=colon]{caption}

\begin{document}

\title{SUAVE: Unified Video-Action Models via Masked Diffusion}

\author{Anonymous authors}

\IEEEaftertitletext{\vspace{-1.2\baselineskip}}



\newif\ifanon
\anonfalse

\ifanon
  \author{Anonymous Authors}
\else
  \author{Rhythm Syed$^{1,2}$, Jean Mercat$^{2}$, Sedrick Keh$^{2}$, Kushal Arora$^{2}$,\\
          Paarth Shah$^{2}$, Aykut Onol$^{2}$, Mengchao Zhang$^{2}$, Tony Dear$^{1}$%
  \thanks{$^{1}$Columbia University, New York, NY, USA.}%
  \thanks{$^{2}$Toyota Research Institute, Los Altos, CA, USA.}%
  \thanks{Corresponding author: Rhythm Syed (\texttt{rhythm.syed@columbia.edu})}}
\fi

\maketitle

  \setlength{\stripsep}{0pt}
  \begin{strip}
      \vspace*{-4.3\baselineskip}
      \centering
      \includegraphics[width=\textwidth]{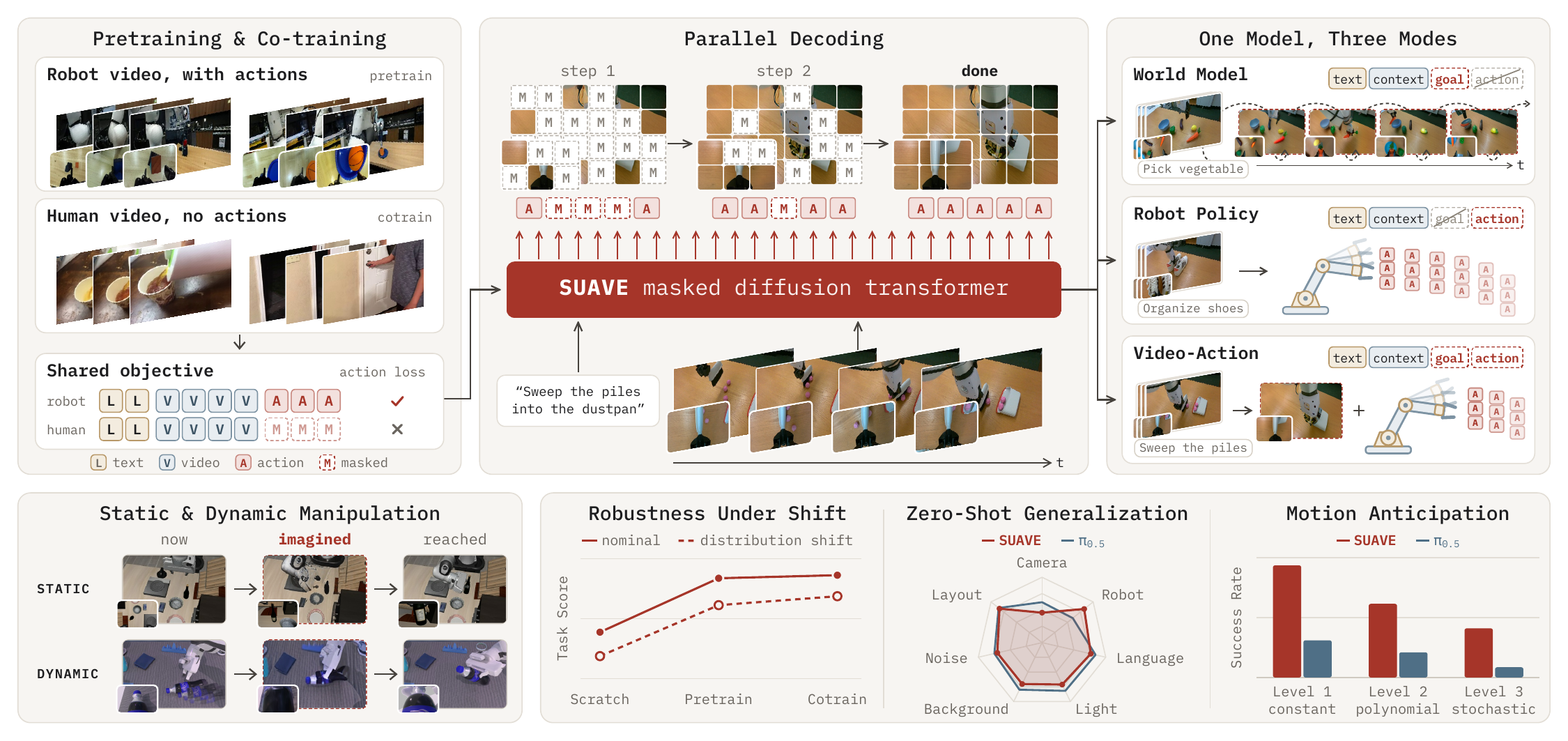}
      \captionof{figure}{\textbf{SUAVE unifies world modeling and control in a single masked diffusion transformer.} \emph{Top:} Robot video with actions and human video with no actions train one model over a shared objective of language, video, and action tokens. Choosing which tokens to mask turns the same model into a world model, a robot policy, or a video-action model. \emph{Bottom:} SUAVE can imagine a goal and then reach it, for various static and dynamic object scenes. Across our evaluations, SUAVE stays robust under distribution shift, generalizes zero-shot to unseen perturbations, and outperforms baselines for motion anticipation.}
      \label{fig:teaser}
      \vspace{10pt} 
  \end{strip}
\begin{abstract}
    Vision-language-action models (VLAs) inherit strong semantic grounding from pretrained vision-language backbones but are typically optimized for predicting actions rather than future observations. They can see and act, but they do not imagine the future before acting. World action models (WAMs) built on video diffusion backbones can imagine but treat language as frozen conditioning on a continuous latent space. Unified models bring these modalities into one architecture, but they either decode autoregressively, one token at a time, or keep video continuous with an auxiliary action head. In this work, we present SUAVE, a Single vocabulary Unified Action-Video modEl in which a masked diffusion transformer generates video and actions conditioned on language, with all three modalities represented as discrete tokens in a shared sequence. Choosing which tokens to mask at inference turns the same network into a world model, a robot policy, or a video-action model. For action-free co-training, the action positions of unlabeled video are filled with mask tokens and excluded from the loss. Simulation and real-world experiments demonstrate two findings. First, a single SUAVE model predicts long-horizon video and acts as a policy, competitive with dedicated world models and specialized action policies on static and dynamic manipulation tasks. On a real robot, our model generates subgoal images and an action chunk spanning one second of motion in 1,030 ms on an RTX 5090 GPU, sustaining closed-loop control at 2.5 actions per second. Second, pretraining on robot video and co-training on human video substantially improves policy performance and zero-shot robustness to distribution shift. Together, these results show that masked diffusion is a practical and versatile foundation for unified video-action modeling.
\end{abstract}


\section{Introduction}

Robot learning has recently experienced a paradigm shift towards end-to-end policies trained on large datasets. This shift has been driven by two complementary model families. Vision-language-action models (VLAs)~\cite{kim2024openvla, black2024pi_0, intelligence2025pi_} adapt pretrained vision-language backbones to output actions and inherit strong semantic grounding. World action models (WAMs)~\cite{ye2026world} adapt pretrained video diffusion backbones to output actions alongside predicted frames and inherit an understanding of physical dynamics. Both families have shown that a pretrained backbone plus robot actions is a viable recipe for general-purpose control. However, each family imposes architectural assumptions that limit its suitability for robotics.

A VLA sees and acts, but it is not trained to anticipate. It maps the current observation to an action chunk without explicitly predicting the visual consequences. Because the standard VLA objective is defined only over actions, it cannot learn from video that lack action labels, which represent a large portion of all video that exists. A WAM is trained to anticipate as a product of its future video prediction, but it does so in a continuous latent space. Although video prediction and control are often jointly trained, many architectures retain separate representations or prediction heads for the two outputs, making them inherently decoupled. Recent work on visual chain of thought~\cite{zhao2025cot} shows that predicting a subgoal image before acting improves VLA performance, which suggests that the two families are converging on the same idea: a policy should be able to imagine the future before acting. This motivates
a single generative model over language, video, and actions, rather than separate modules for each.

We model the three modalities in a shared sequence using masked diffusion, a method that is analogous to continuous diffusion but in the discrete space. If every video frame, action, and language token is drawn from one vocabulary, then a transformer trained to unmask tokens can learn to fill in video and actions across a few denoising steps. Which conditional the model samples at inference time is decided entirely by which positions are masked: mask the future frame and it becomes a world model, mask the actions and it becomes a robot policy, mask both and it becomes a video-action model that generates a visual subgoal and the actions needed to reach the subgoal. Excluding masked action positions from the loss allows video without action labels to enter the same batch as labeled demonstrations, enabling robot video with actions and human video without actions to share one training stage and one unified objective. Because all masked positions are predicted in parallel, decoding at inference only takes a handful of steps rather than one step per token. This is in contrast to fully autoregressive methods, which make generating a subgoal image alongside actions prohibitively slow for closed-loop control, as each image and action token needs to be predicted sequentially. The discrete formulation enables strong versatility, allowing the same model weights to control the robot, roll out long-horizon video, and learn from human video all in one objective.

We propose SUAVE, a \textbf{S}ingle-vocabulary \textbf{U}nified \textbf{A}ction-\textbf{V}ideo mod\textbf{E}l built on a masked diffusion transformer. SUAVE arranges a video context, a language instruction, a predicted goal frame, and an action chunk as one token sequence. Because every token is predicted by the same objective, the mask at inference selects whether the model acts as a world model, a robot policy, or a video-action model. Experiments in simulation and on a real robot support two findings:

\begin{enumerate}
    \item \textbf{One model is a competitive world model and robot policy.} A single SUAVE checkpoint predicts long-horizon video at a quality comparable to dedicated world models. As a policy, it is competitive with specialized action policies on static manipulation tasks and exceeds every baseline on dynamic manipulation tasks. We show our results for real world tasks, LIBERO, LIBERO-PLUS, and the DOMINO benchmark.
  \item \textbf{Pretraining and co-training on video substantially improve performance and zero-shot robustness.} Pretraining on robot video and co-training on action-free human video more than double the real robot task scores of models trained from scratch and roughly triple them under distribution shift, with the gap statistically significant in nearly every task and condition.
\end{enumerate}
\section{Related Work}

\subsection{Vision-language-action models}
VLAs adapt a pretrained vision-language backbone to robot control by treating actions as an additional output modality. RT-2~\cite{brohan2023rt} and OpenVLA~\cite{kim2024openvla} discretize actions into tokens and decode them autoregressively. OpenVLA-OFT~\cite{kim2025fine} replaces sequential decoding with parallel regression over an action chunk. $\pi_0$~\cite{black2024pi_0}, $\pi_0$-FAST~\cite{pertsch2025fast}, and $\pi_{0.5}$~\cite{intelligence2025pi_} attach flow-matching heads or compressed action tokenizers to pretrained VLM backbones. All of these map the current observation and language instruction to action. CoT-VLA~\cite{zhao2025cot} extends this by predicting a single subgoal image before acting and shows that this visual chain of thought approach improves success through better grounding. However, because the subgoal is generated token by token, the authors report a 7$\times$ slowdown in control. These models generate actions and at most one frame, however none model multi-step video. As their objectives are defined over actions, none can train on video that lacks action labels.

\subsection{World models and world action models}
A parallel line of work starts from video generation. iVideoGPT~\cite{wu2024ivideogpt}, Interactive World Simulator~\cite{wang2026interactive}, and Cosmos~\cite{agarwal2025cosmos} learn action- or latent action-conditioned video predictors. Recent world action models, such as DreamZero~\cite{ye2026world}, adapt pretrained video diffusion backbones to output actions alongside predicted frames. These models can imagine the future well, but they typically operate in a continuous latent space: when language is present, it enters only as conditioning from a frozen text encoder, and actions are produced by a separate head. Video, actions, and language are therefore handled by different components with different objectives. SUAVE takes the opposite approach: all three are represented as discrete tokens and processed by one network under a shared objective.

\subsection{Unified video-action and unified multimodal models}
Several recent works generate video and actions jointly. UWM~\cite{zhu2025unified} and UVA~\cite{li2025unified} are continuous diffusion models over frames and actions. UWM samples independent noise levels for the two modalities so that one network can act as a policy, a forward model, or an inverse dynamics model. It is the closest continuous analog to our approach, and our independent per-modality mask rates are the discrete counterpart of its independent time steps. UniVLA~\cite{wang2026unified} unifies vision, language, and action as discrete tokens but decodes them autoregressively, one token at a time. On the language-model side, LLaDA~\cite{nie2026large} showed that masked diffusion scales to LLM-sized text models, and MMaDA~\cite{yang2026mmada} extends it to a shared text and image vocabulary with parallel decoding. SUAVE brings this backbone to robotics by adding video and actions. Existing designs fall into three families: models that decode autoregressively over discrete tokens, models with continuous diffusion over frames and actions, and hybrid models that combine a language model with diffusion heads. SUAVE presents a fourth design: one generative mechanism over one discrete vocabulary for every modality, so that all cross-modal interaction happens under a single objective, with subgoal frames and the action chunk decoded in parallel.

\subsection{Pretraining and co-training on video for robot policies}
Large scale video is a promising source of transferable supervision for robotics. Prior works have found several ways to use it in training. UniVLA~\cite{wang2026unified} post-trains on video before learning a policy. UWM~\cite{zhu2025unified} sets the action diffusion timestep of action-free clips to maximum noise, allowing them to train alongside labeled demonstrations. SUAVE achieves this in the discrete setting by filling action positions with mask tokens and excluding them from the loss, so a human video clip and a robot demonstration share a batch. Our evaluation methodology follows LBM~\cite{barreiros2025careful}, which shows controlled comparisons of data recipes on a fixed architecture with quantified uncertainty. We evaluate on LIBERO~\cite{liu2023libero}, LIBERO-Plus~\cite{fei2025libero} for robustness to perturbations, and DOMINO~\cite{fang2026towards} for dynamic manipulation. Our contribution on this axis is a controlled evaluation of the effects of robot-video pretraining and action-free human video co-training, conducted under one protocol in simulation and on a real robot.
\begin{figure*}[t]
      \centering
      \includegraphics[width=\textwidth]{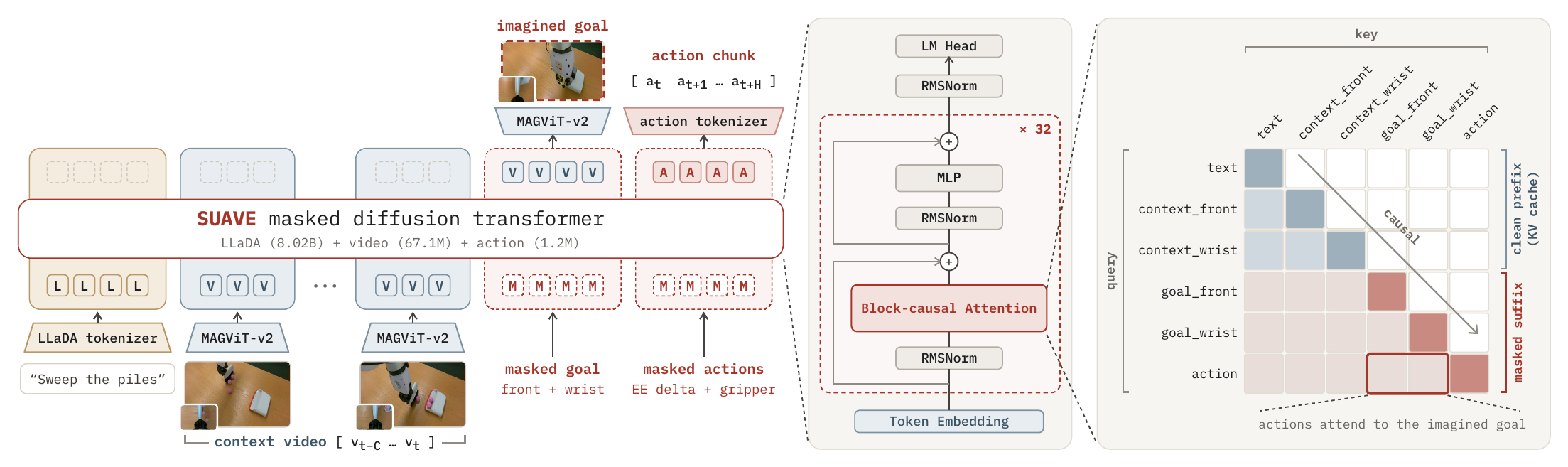}
      \caption{\textbf{Architecture.} \emph{Left:} The instruction is tokenized by the LLaDA tokenizer, and the video context is encoded by a MAGViT-v2 tokenizer. The goal and action tokens start fully masked, and a masked diffusion transformer, initialized from LLaDA with MMaDA's image-token embeddings plus a new action vocabulary, denoises them into the imagined goal frame and the action chunk. \emph{Middle:} Each of the 32 layers is a standard LLaDA block whose self-attention uses a block-causal mask. \emph{Right:} Each cell in the mask is one block. Attention is bidirectional within a block and causal across blocks. The clean prefix is computed once and cached, and only the masked suffix is recomputed at each denoising step.}
    \label{fig:method}
  \end{figure*}
\section{Method}

SUAVE is a single transformer that models language, video, and robot actions as one sequence of discrete tokens, as shown in Figure~\ref{fig:method}. Attention is bidirectional within each block of the sequence and causal across blocks, and the model is trained with a masked diffusion objective~\cite{austin2021structured, sahoo2024simple, shi2024simplified}. In this section, we review masked diffusion (\S\ref{sec:background}), describe how each modality is sequenced (\S\ref{sec:sequence}), state the training objective (\S\ref{sec:objective}), and explain how the model is decoded as a world model, a policy, and a video-action model (\S\ref{sec:inference}).

\subsection{Background}
    \label{sec:background}
    
Let $\mathbf{x}_0 = (x_0^1,\dots,x_0^N)$ be a sequence of tokens from a vocabulary $\mathcal{V}$ augmented with a mask token $[\mathrm{M}]$. Masked diffusion~\cite{austin2021structured, chang2022maskgit, nie2026large} defines a forward process
that progressively replaces tokens with $[\mathrm{M}]$ and a reverse process that restores them. In the forward process, each token of $\mathbf{x}_0$ is masked independently with probability $t$, where $t\in(0,1]$ is the noise level, resulting in a corrupted sequence $\mathbf{x}_t$. Because the corruption process replaces a token with
$[\mathrm{M}]$, $\mathbf{x}_t$ contains a mixture of clean tokens and mask tokens. The mask state is absorbing, where a token that is masked at level $t$ remains masked at every higher level, and at $t=1$ every token is masked. The reverse process runs from $t=1$ to $t=0$ and is driven by a mask predictor $p_\theta(x_0^i \mid \mathbf{x}_t)$, a network that predicts the clean token at every masked position from the partially masked sequence. The generation process alternates between predicting mask tokens and re-masking a fraction of them at the next noise level until no mask tokens remain. The mask predictor is trained with cross-entropy on the masked positions, which we state for SUAVE in \S\ref{sec:objective}.

Two properties make this objective suitable for robotics. First, the mask predictor predicts every token in parallel in a single forward pass, where a full sequence is generated in a small number of denoising steps rather than one step per token. Second, the model can denoise a sequence with any combination of mask tokens, so the conditional it samples at inference is chosen by which tokens are masked. LLaDA~\cite{nie2026large} trained this objective and released open weights, while industry language model systems such as Gemini Diffusion~\cite{deepmind2025gemini} and Mercury~\cite{inception2025mercury} remain closed. MMaDA~\cite{yang2026mmada} extends this objective to a shared text and image vocabulary, and SUAVE extends that to video and robot actions.

\subsection{Tokenization and Sequence Layout}
    \label{sec:sequence}

\textbf{Backbone.} SUAVE is initialized from MMaDA-8B~\cite{yang2026mmada}, a 32-layer transformer built on the LLaDA architecture~\cite{nie2026large}. We make two changes. The attention mask is replaced by the block-causal mask described below, and the vocabulary is extended with 256 action tokens placed directly after the vision codebook. No modality-specific encoders, adapters, or action heads are added.

\textbf{Tokenizers.} Every modality is mapped into one vocabulary composed of three disjoint sets, $\mathcal{V} = \mathcal{V}_{\text{text}} \cup \mathcal{V}_{\text{video}} \cup \mathcal{V}_{\text{action}}$, that share a single embedding table and output head. \emph{Language} uses the LLaDA text tokenizer. \emph{Video} is encoded by MAGViT-v2~\cite{yu2024language}, a convolutional image tokenizer with lookup-free quantization. With this, a $256\times256$ pixel image becomes a $16\times16$ grid of tokens, where each frame occupies $256$ tokens. The tokenizer is frozen and shared by all datasets. \emph{Actions} are tokenized per dimension with uniform bins~\cite{brohan2023rt, kim2024openvla}. Each dimension is linearly rescaled to $[-1,1]$ using its q1 and q99 quantiles in the training dataset, clipped to that range, and assigned to one of $256$ bins. All dimensions and time steps share these 256 ids and are distinguished by their position in the action block. At inference, decoding returns the bin center.

\textbf{Sequence.} A training sample of our token sequence is shown in Eq.~\eqref{eq:sequence} read from left to right and top to bottom:
\newcommand{\stok}[1]{\mathtt{[#1]}}
\newcommand{\sblk}[1]{\{\mathrm{#1}\}}
  \begin{equation}
  \small
  \begin{aligned}
    &\stok{BOS}\,\sblk{text}\,\stok{EOS} \\[2pt]
    &\underbrace{\stok{SOI}\,\sblk{front}_k\,\stok{EOI}\;
                 \stok{SOI}\,\sblk{wrist}_k\,\stok{EOI}}_{k=1,\dots,8\;\;\mathrm{(video\ context)}} \\[2pt]
    &\underbrace{\underbrace{\stok{SOI}\,\sblk{goal}^{f}\,\stok{EOI}\;
                             \stok{SOI}\,\sblk{goal}^{w}\,\stok{EOI}}_{\mathrm{world\ model}}
                 \;
                 \underbrace{\stok{SOA}\,\sblk{act}\,\stok{EOA}}_{\mathrm{policy}}}_{\mathrm{video\text{-}action\ model}}
  \end{aligned}
  \label{eq:sequence}
  \end{equation}
Here $\stok{BOS}$ and $\stok{EOS}$ are the beginning and end of sentence delimiter tokens of the language tokenizer. $\sblk{text}$ is the language instruction. $\stok{SOI}$ and $\stok{EOI}$ mark the start and end of one image. $\stok{SOA}$ and $\stok{EOA}$ mark the start and end of an action chunk. $\sblk{front}_k$ and $\sblk{wrist}_k$ are the $256$ tokens of the front and wrist frames, each at the $k$-th context step. $\sblk{goal}^{f}$ and $\sblk{goal}^{w}$ are the two goal images at the end of the action chunk. $\sblk{act}$ is the $K \times D$ action tokens, where $K$ is the number of consecutive future actions in the chunk, and $D$ is the action dimension of the robot. The braces name the mode obtained by masking that block at inference: masking the goal frames gives a world model, masking the action chunk gives a policy, and masking both together gives a video-action model.

\textbf{Block-causal attention.} Every token carries an index within a block, where we have one block for the instruction, one per video frame, and one for the action chunk, in sequential order. Attention is bidirectional within a block and causal across blocks, as shown in Figure~\ref{fig:method}. The goal frames attend to the context and language instruction, and the action chunk attends to the goal frames. The policy is therefore goal-conditioned, where actions depend on the imagined future. Within a video frame, all $256$ tokens attend to one another.

\subsection{Training Objective}
    \label{sec:objective}

For joint video-action prediction, the targets are the goal-frame tokens and the action tokens. Following UWM~\cite{zhu2025unified}, which samples independent diffusion time steps for its observation and action streams, we draw independent mask rates for the video and action modalities ($r_v$, $r_a$). For each training sequence of the form in Eq.~\eqref{eq:sequence}, we sample $u_v, u_a \sim \mathcal{U}(0,1)$,
\begin{equation}
  r_v = \cos\!\big(\tfrac{\pi}{2}u_v\big), \qquad r_a = \cos\!\big(\tfrac{\pi}{2}u_a\big),
\end{equation}
and mask a uniformly random fraction $r_v$ of the $512$ goal-frame tokens ($256$ $\times$ 2 (front and wrist view)) and $r_a$ of the $K \times D$ action tokens. This is the fixed fraction corruption of MaskGIT~\cite{chang2022maskgit} with a cosine schedule.

Let $\mathcal{P}$ be the set of supervised positions, the masked target tokens that carry a label. The loss is the mean cross-entropy over $\mathcal{P}$,
  \begin{equation}
    \mathcal{L}(\theta) = -\,\mathbb{E}_{\mathbf{x}_0,\,\tilde{\mathbf{x}}}\Big[\frac{1}{|\mathcal{P}|}\sum_{i\in\mathcal{P}} \log p_\theta\big(x_0^i \mid \tilde{\mathbf{x}}\big)\Big],
    \label{eq:loss}
  \end{equation}
where $\tilde{\mathbf{x}}$ is the sequence with the selected positions masked out and the expectation is over training sequences and mask draws. Target tokens not selected for masking remain visible and are excluded from the loss. Because $r_v$ and $r_a$ are drawn independently, training covers every combination of masked targets that is used during inference.

\textbf{Action-free video.} A demonstration without action labels, such as a clip of human video, is laid out as shown in Eq.~\eqref{eq:sequence}, with the action block padded with mask tokens. Those positions have no labels, so they are not in $\mathcal{P}$, and the example is supervised only through its goal frames. Its context frames and language instruction are still attended to. This is our co-training mechanism: action-labeled robot demonstrations and action-free human video share a batch, a loss, and a training stage, with no auxiliary objective.
    \label{sec:action-free}

\subsection{Inference}
\label{sec:inference}

\textbf{Modes.} At inference time, the context video and the language instruction are given clean, and the target blocks are masked according to the desired mode. All policy results in this work use the video-action mode. In closed-loop control, the generated goal frame is discarded, and the next context frame comes directly from the camera observation.

\textbf{Long-horizon video rollout.} Video beyond one goal frame is generated block-autoregressively, where the decoded goal frame becomes the newest context and the oldest context frame is dropped. Each step advances one second of video, and in our experiments, the rollout continues until the end of the held-out episode, up to 40 frames on the real tasks. Rollouts are seeded with a cold-start from a single observed frame, repeated across all eight context steps. During training, the same rule is applied to context indices that fall before the start of an episode.

\textbf{Prefix key-value cache.} Under the block-causal mask, tokens preceding the first masked position never attend to other masked tokens, so their keys and values are identical across the denoising steps. They are computed once per action chunk and reused, and each denoising step runs a forward pass over the masked suffix only. The cache is exact, producing bit-identical outputs, and reduces decode time per chunk by up to $2.2\times$. With 4 denoising steps, which we use across all of our experiments, the video-action mode generates front and wrist subgoal images and a chunk of five actions spanning one second of motion in 1,030 ms on an RTX 5090 GPU, sustaining closed-loop control at about 2.5 actions per second on a real xArm7 robot.

\section{Experiments}
    \label{sec:experiments}

Our experiments investigate three questions. (Q1) Is SUAVE competitive with specialized action policies in various static and dynamic manipulation scenes? (Q2) How does SUAVE's video prediction compare with dedicated world models trained on the same data? (Q3) What do pretraining on robot video and co-training on action-free human video each contribute in the discrete space formulation under nominal conditions and distribution shift? We evaluate on a 7-DoF xArm7 robot for real tasks and on the LIBERO, LIBERO-Plus, and DOMINO benchmarks for simulation.

\subsection{Experimental Setup}
    \label{sec:setup}

\textbf{Model variants.} All SUAVE models share the same architecture and hyperparameters, differing only in training data. \emph{SUAVE-Scratch} initializes from MMaDA-8B and trains directly on the target benchmark's demonstrations. \emph{SUAVE-Pretrain} first trains on a robot video corpus with actions, then finetunes on the target demonstrations. \emph{SUAVE-Cotrain} pretrains on the same robot corpus mixed one-to-one with action-free human video from Something-Something v2 (SSv2)~\cite{goyal2017something}, then finetunes in the same manner. Co-training here refers to a single pretraining stage where labeled and unlabeled videos share batches, as explained in Section~\eqref{sec:action-free}. The robot corpus is chosen per benchmark so that pretraining stays in-domain. For real world tasks, we pretrain on DROID~\cite{khazatsky2024droid}. For LIBERO and LIBERO-Plus, we pretrain LIBERO-90, the 90 tasks that are disjoint from the four evaluation suites. For DOMINO, we pretrain on non-Franka embodiments, where evaluation only uses the Franka arm tasks. Each model is trained with a single seed, and we use the final checkpoint. Every policy decodes an action chunk of five actions per query at 5 Hz with 4 denoising steps and a prefix KV cache.

\begin{figure*}[!t]
  \centering
  \includegraphics[width=\linewidth]{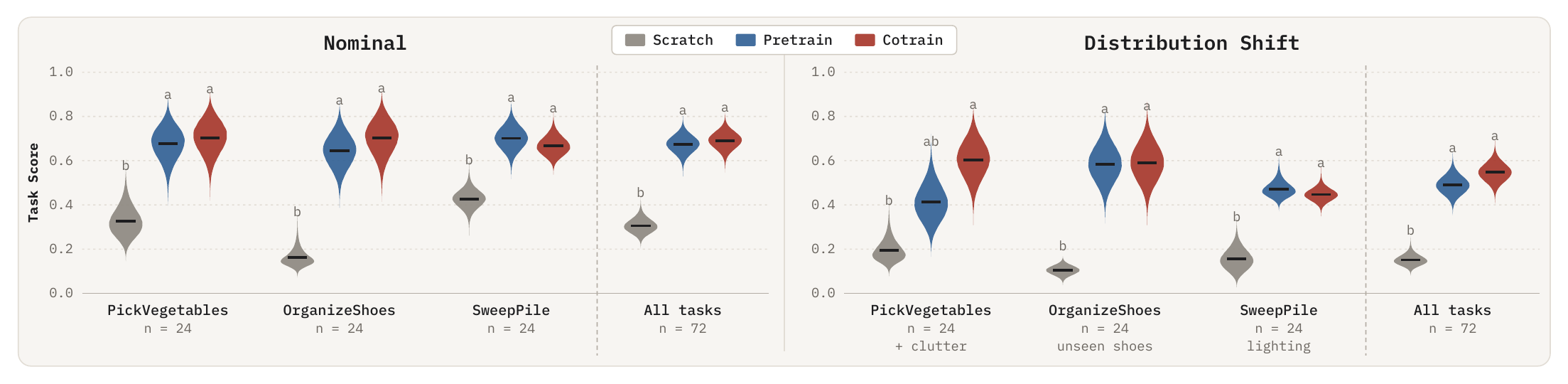}
  \caption{\textbf{Real robot task score under nominal conditions and distribution shift.} Each violin is the posterior of the mean Task Score for one model variant on one task, computed by a Bayesian bootstrap over its 24 rollouts. The black bar is the mean. The left half shows nominal conditions and the right half the distribution shift with added clutter, unseen shoes, altered lighting. Models labeled with different letters are statistically distinguishable.}
  \label{fig:realtasks}
\end{figure*}

\textbf{Real Setup.} With the xArm7 robot, we use a wrist-mounted RealSense D405 and an over-the-shoulder front view with a RealSense D455. We collect three tasks with 100 demonstrations each. \emph{PickVegetables}: pick the vegetable named in the instruction from four candidates, lemon, green pepper, eggplant, or carrot. \emph{OrganizeShoes}: correctly orient a fallen shoe and align it with the other shoe. \emph{SweepPile}: sweep three debris piles into a dustpan with a brush held in the gripper. For each task, we train the three model variants and evaluate each on 24 initial
configurations under nominal conditions, and then on the same 24 configurations under a task-specific distribution shift, which includes novel distractor objects for PickVegetables, an unseen shoe for OrganizeShoes, and altered lighting for SweepPile. Each episode receives a \emph{Task Score}, defined as the unweighted mean of ten components assigned by a reviewer who is blind to the model and condition. The components measure progress in engagement, attempts, partial completion, collateral contact, timeout, and speed. Specific component definitions are task-specific and are released with the code. Following LBM~\cite{barreiros2025careful}, we report the posterior of the mean task score under a Bayesian bootstrap with a uniform Dirichlet prior over the observed scores and assign a shared letter to models whose means are not separated by a pairwise Welch $t$-test at $\alpha$=$0.05$ with Bonferroni correction.

\begin{figure*}
    \centering
    \includegraphics[width=\linewidth]{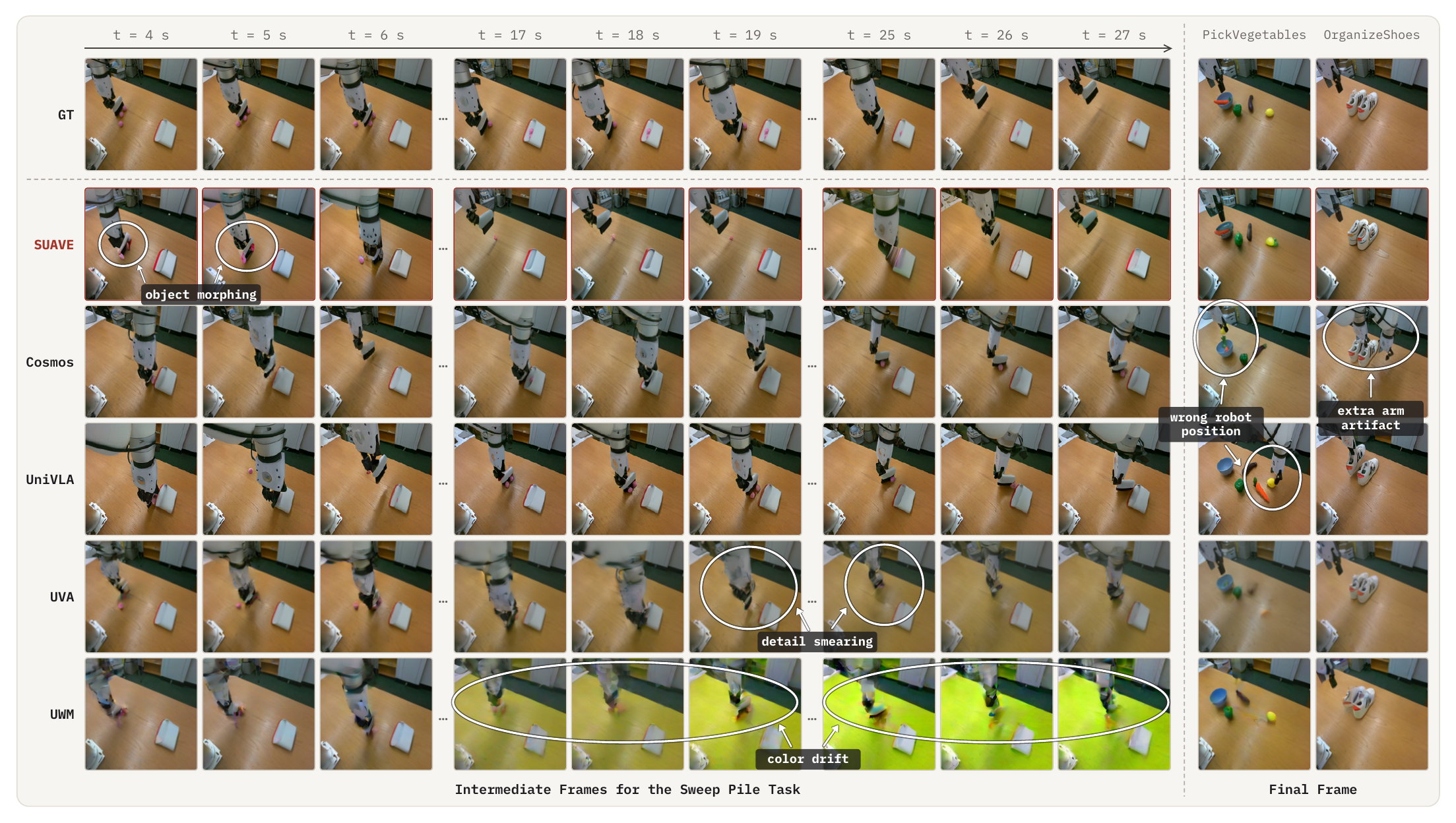}
    \caption{\small \textbf{Qualitative Comparison of Long-Horizon Video Prediction.}
    Rows show the ground-truth rollout (GT), SUAVE (red), and four baselines.
    Left: predicted frames for the Sweep Pile task at the indicated times (in seconds).
    Right: final predicted frame for Pick Vegetables and Organize Shoes.
    White circles mark failure modes.}
    \label{fig:quali}
\end{figure*}
  \begin{table}[!t]
  \caption{\textbf{Video prediction on real tasks.} Rollouts from the first observed frame to the end of the episode on 8 held-out episodes per task. All predicted frames of a task are pooled against the raw $256{\times}256$ camera frame. Baselines are trained by us with the authors' released recipes and weights on the same demonstrations. SUAVE rows are the same checkpoints of Fig.~\ref{fig:realtasks}.}
  \label{tab:wm_metrics}
  \centering
  \small
  \setlength{\tabcolsep}{2.5pt}
  \begin{tabular}{lcccc}
    \toprule
    Method & PSNR $\uparrow$ & SSIM $\uparrow$ & LPIPS $\downarrow$ & FVD $\downarrow$ \\
    \midrule
    Cosmos~\cite{agarwal2025cosmos}          & \pmv{17.8}{1.2} & \pmv{\textbf{0.61}}{0.04} & \pmv{\textbf{0.22}}{0.04} & \pmv{\textbf{472}}{33} \\
    iVideoGPT~\cite{wu2024ivideogpt}       & \pmv{17.2}{2.3} & \pmv{0.39}{0.12} & \pmv{0.46}{0.15} & \pmv{2939}{1025} \\
    UVA~\cite{li2025unified}             & \pmv{\textbf{19.3}}{1.7} & \pmv{0.60}{0.06} & \pmv{0.33}{0.04} & \pmv{2368}{1356} \\
    UWM~\cite{zhu2025unified}             & \pmv{15.9}{1.0} & \pmv{0.38}{0.05} & \pmv{0.42}{0.09} & \pmv{1922}{524} \\
    UniVLA~\cite{wang2026unified}          & \pmv{17.9}{1.9} & \pmv{0.58}{0.05} & \pmv{0.27}{0.06} & \pmv{1095}{195} \\
    \midrule
    SUAVE-Scratch   & \pmv{18.4}{1.4} & \pmv{0.55}{0.04} & \pmv{0.23}{0.05} & \pmv{1883}{588} \\
    SUAVE-Pretrain  & \pmv{18.3}{1.3} & \pmv{0.55}{0.04} & \pmv{0.23}{0.05} & \pmv{2005}{558} \\
    SUAVE-Cotrain   & \pmv{18.4}{1.3} & \pmv{0.55}{0.03} & \pmv{0.23}{0.05} & \pmv{1709}{400} \\
    \bottomrule
  \end{tabular}
  \end{table}

\textbf{Discussion.} Figure~\ref{fig:realtasks} shows the posterior of the mean task score for each variant, task, and condition. Under nominal conditions, the pooled task score increases from 0.31 for Scratch to 0.67 for Pretrain and 0.69 for Cotrain. Under distribution shift, Scratch falls to 0.15, while Pretrain and Cotrain hold at 0.49 and 0.55, respectively. The pretrained variants lose a smaller fraction of their nominal score and retain a higher absolute score. Scratch receives its own letter in every column except PickVegetables under shift, where Pretrain shares it, showing that the gap between Scratch and the pretrained variants is statistically significant in every other task and condition. Pretrain and Cotrain share a letter in every column, showing that the additional effect of human-video co-training is not statistically distinguishable at this sample size.

\textbf{Video prediction.} The same checkpoints also serve as world models for video rollouts. Table~\ref{tab:wm_metrics} compares the rollouts of SUAVE on eight held-out episodes per task against dedicated video predictors (Cosmos~\cite{agarwal2025cosmos} and iVideoGPT~\cite{wu2024ivideogpt}) and unified multimodal models (UVA~\cite{li2025unified}, UniVLA~\cite{wang2026unified}, UWM~\cite{zhu2025unified}), all trained on the same demonstrations and scored against the raw camera frames, with metrics pooled over all predicted frames of a task and then averaged. SUAVE, across all its variants, performs comparably to Cosmos, the strongest video model among the baselines, by exceeding it in PSNR and approaching it in LPIPS, but is lower in SSIM and substantially worse in FVD. Cosmos is a 2B-parameter model trained only to predict video at ten frames per second. SUAVE's advantage is its versatility, as a single set of weights serves many capabilities, including video prediction. Against the unified models, SUAVE achieves the best LPIPS and outperforms UWM on PSNR, SSIM, and LPIPS, with comparable FVD. UVA leads in PSNR and SSIM but is worse in LPIPS and FVD, while UniVLA attains higher SSIM and lower FVD than SUAVE. The three SUAVE variants are indistinguishable in PSNR, SSIM, and LPIPS, however, co-training gives the lowest FVD among them. Fig.~\ref{fig:quali} shows the qualitative comparison of these rollouts. SUAVE maintains scene structure and reaches final states that closely match the ground truth, with minor object morphing early in the rollout due to reconstruction artifacts from the video tokenizer, while the baselines exhibit occasional artifacts such as incorrect robot positions, hallucinated artifacts, and color drift.

  \begin{table}[!t]
  \caption{\textbf{LIBERO success rates} (\%, higher is better). Mean and standard error over 500 trials per suite. $^{*}$UWM is retrained by us as a single multitask language-conditioned policy. Bold
  marks the best per column.}
  \label{tab:full-libero}
  \centering
  \small
  \setlength{\tabcolsep}{2.5pt}
  \begin{tabular}{lccccc}
    \toprule
    Method & Spatial & Object & Goal & Long & Avg \\
    \midrule
    DP~\cite{chi2025diffusion}              & \pmv{78.3}{1.1} & \pmv{92.5}{0.7} & \pmv{68.3}{1.2} & \pmv{50.5}{1.3} & \pmv{72.4}{0.7} \\
    OpenVLA~\cite{kim2024openvla}         & \pmv{84.7}{0.9} & \pmv{88.4}{0.8} & \pmv{79.2}{1.0} & \pmv{53.7}{1.3} & \pmv{76.5}{0.6} \\
    CoT-VLA~\cite{zhao2025cot}         & \pmv{87.5}{1.4} & \pmv{91.6}{0.5} & \pmv{87.6}{0.6} & \pmv{69.0}{0.8} & \pmv{81.1}{0.6} \\
    UWM$^{*}$~\cite{zhu2025unified}       & \pmv{84.0}{1.6} & \pmv{84.4}{1.6} & \pmv{87.4}{1.5} & \pmv{84.0}{1.6} & \pmv{85.0}{0.8} \\
    UniVLA~\cite{wang2026unified}          & \pmv{94.2}{1.0} & \pmv{97.2}{0.7} & \pmv{92.4}{1.2} & \pmv{91.8}{1.2} & \pmv{93.9}{0.5} \\
    $\pi_{0.5}$~\cite{intelligence2025pi_}     & \pmv{\textbf{98.6}}{0.5} & \pmv{99.0}{0.4} & \pmv{\textbf{98.6}}{0.5} & \pmv{93.2}{1.1} & \pmv{\textbf{97.4}}{0.4} \\
    \midrule
    SUAVE-Scratch   & \pmv{87.8}{1.5} & \pmv{96.0}{0.9} & \pmv{93.0}{1.1} & \pmv{94.8}{1.0} & \pmv{92.9}{0.6} \\
    SUAVE-Cotrain   & \pmv{90.6}{1.3} & \pmv{\textbf{99.4}}{0.3} & \pmv{98.0}{0.6} & \pmv{\textbf{95.4}}{0.9} & \pmv{95.9}{0.4} \\
    \bottomrule
  \end{tabular}
  \end{table}

\subsection{LIBERO and LIBERO-Plus for Static Manipulation}
  \label{sec:libero}

\textbf{Setup.} LIBERO~\cite{liu2023libero} has four suites of ten tasks each: Spatial, Object, Goal, and Long, with 50 demonstrations per task. SUAVE is trained on all four suites and evaluated for 50 trials per task. LIBERO-Plus~\cite{fei2025libero} perturbs the 40 tasks along seven axes: camera viewpoint, robot initial state, language, lighting, background texture, sensor noise, and object layout, resulting in 10,030 variants evaluated once each. We report a \emph{zero-shot} setting with models trained only on clean demonstrations from the original LIBERO suite and an \emph{in-distribution} setting with models finetuned on the LIBERO-Plus training mix. As an ablation to separate in-domain robot data from out-of-domain data, we also train a five-rung pretraining ladder on LIBERO-Long alone, with four runs per rung. We evaluate each rung on clean LIBERO-Long and zero-shot on its perturbed variants to understand the contribution of each recipe.

\begin{table}[!t]
    \caption{\textbf{Pretraining ladder on LIBERO-Long (\%):} All rows finetune the same model on LIBERO-Long and differ only in pretraining corpus. We show the original LIBERO mean and standard error over 4 seeds $\times$ 500 trials each and the LIBERO-Plus zero-shot success on 2,519 trials each.}
  \centering
  \small
  \setlength{\tabcolsep}{3pt}
  \begin{tabular}{llcc}
      \toprule
      & & \multicolumn{2}{c}{Long horizon suite} \\
      \cmidrule(lr){3-4}
      Method & Pretraining corpus & LIBERO & LIBERO-Plus \\
      & & & (zero-shot) \\
      \midrule
      SUAVE-Scratch  & none             & \pmv{87.7}{0.7} & \pmv{52.1}{1.0} \\
      SUAVE-Pretrain & DROID            & \pmv{87.2}{0.7} & \pmv{53.1}{1.0} \\
      SUAVE-Cotrain  & DROID + SSv2     & \pmv{88.7}{0.7} & \pmv{53.9}{1.0} \\
      SUAVE-Pretrain & LIBERO-90        & \pmv{93.6}{0.5} & \pmv{63.4}{1.0} \\
      SUAVE-Cotrain  & LIBERO-90 + SSv2 & \pmv{\textbf{94.5}}{0.5} & \pmv{\textbf{63.8}}{1.0} \\
      \bottomrule
  \end{tabular}
  \label{tab:libero_ladder}
\end{table}

\textbf{Discussion.} Table~\ref{tab:full-libero} answers Q1 for static scenes in the original LIBERO suite. SUAVE-Cotrain reaches 95.9\% average, second to $\pi_{0.5}$ at 97.4\% and ahead of UniVLA at 93.9\%. SUAVE-Cotrain is the best model on Long at 95.4\%. SUAVE-Scratch, with no pretraining, reaches 92.9\%, within one point of UniVLA. UWM, trained on the same data as a single multitask policy, reaches 85.0\%. Table~\ref{tab:libero_ladder} shows the pretraining ladder on LIBERO-Long. The two DROID rows are within standard error of Scratch on the original suite and add at most two points zero-shot on the perturbed suite, showing that out-of-domain robot data transfers little to these simulation tasks. The two LIBERO-90 rows add six points in LIBERO and eleven points in LIBERO-Plus. Adding SSv2 on top of LIBERO-90 raises the original score by 0.9 points and the zero-shot score by 0.4 points. The results suggest that the observed gains are primarily associated with in-domain action-labeled data, with a small measurable contribution from human video at this scale. Guided by this result, the Cotrain variants in the remaining experiments train on in-domain robot data.

Table~\ref{tab:liberoplus} answers Q3 on robustness. In the zero-shot setting, SUAVE-Cotrain reaches 71.3\% across all perturbations, above every LIBERO-trained policy except for $\pi_{0.5}$, which we evaluate from its released checkpoint at 84.7\%. SUAVE-Scratch at 61.8\% sits in the middle of the table, level with $\pi_0$-FAST. The two SUAVE models share identical architecture and finetuning data and differ only in the pretraining recipe, to which the 9.5-point gap can be attributed. Camera viewpoint is the hardest axis for both SUAVE models, as it is for most of the baselines. Cotrain is strongest on robot initial state, where its 82.5\% is the highest among the evaluated methods, including $\pi_{0.5}$. In-distribution, finetuning on the perturbed demonstrations brings SUAVE-Cotrain to 86.1\%, outperforming OpenVLA-OFT.

  \begin{table}[!t]
  \caption{\textbf{LIBERO-Plus success rates (\%) by perturbation category.} All rows cover the full benchmark (10{,}030 tasks, one trial each). SUAVE-Cotrain is pretrained on LIBERO-90 and SSv2. Bold marks the best per column.}
  \label{tab:liberoplus}
  \centering
  \small
  \setlength{\tabcolsep}{2.5pt}
  \begin{tabular}{lcccccccc}
      \toprule
      Method & \rotatebox{90}{Camera} & \rotatebox{90}{Robot} & \rotatebox{90}{Language} & \rotatebox{90}{Light} & \rotatebox{90}{Background} & \rotatebox{90}{Noise} & \rotatebox{90}{Layout} & \rotatebox{90}{Total} \\
      \midrule
      \multicolumn{9}{l}{\textit{Zero-shot: trained on LIBERO demonstrations only}} \\
      OpenVLA~\cite{kim2024openvla}              & 0.8  & 3.5  & 23.0 & 8.1  & 34.8 & 15.2 & 28.5 & 15.6 \\
      OpenVLA-OFT~\cite{kim2025fine}          & 56.4 & 31.9 & 79.5 & 88.7 & 93.3 & 75.8 & 74.2 & 69.6 \\
      OpenVLA-OFT$_w$      & 10.4 & 38.7 & 70.5 & 76.8 & 93.6 & 49.9 & 69.9 & 55.8 \\
      OpenVLA-OFT$_m$      & 55.6 & 21.7 & 81.0 & 92.7 & 91.0 & 78.6 & 68.7 & 67.9 \\
      $\pi_0$~\cite{black2024pi_0}              & 13.8 & 6.0  & 58.8 & 85.0 & 81.4 & 79.0 & 68.9 & 53.6 \\
      $\pi_0$-FAST~\cite{pertsch2025fast}         & 65.1 & 21.6 & 61.0 & 73.2 & 73.2 & 74.4 & 68.8 & 61.6 \\
      $\pi_{0.5}$~\cite{intelligence2025pi_}      & \textbf{71.5} & 75.5 & \textbf{86.4} & \textbf{97.5} & \textbf{95.6} & \textbf{86.6} & \textbf{86.6} & \textbf{84.7} \\
      UniVLA~\cite{wang2026unified}  & 52.3 & 39.9 & 70.5 & 84.0 & 64.5 & 81.2 & 78.4 & 66.7 \\
      UWM$^{*}$~\cite{zhu2025unified}            & 4.8  & 22.3 & 27.7 & 36.2 & 25.4 & 6.2  & 36.7 & 21.9 \\
      SUAVE-Scratch        & 29.4 & 73.5 & 70.6 & 66.1 & 59.1 & 56.5 & 79.0 & 61.8 \\
      SUAVE-Cotrain        & 45.9 & \textbf{82.5} & 76.6 & 71.2 & 70.3 & 70.3 & 83.4 & 71.3 \\
      \midrule
      \multicolumn{9}{l}{\textit{In-distribution: trained on LIBERO-Plus perturbed data}} \\
      OpenVLA-OFT$_m$ + PT & 92.8 & 30.3 & \textbf{85.8} & \textbf{94.9} & 93.9 & 89.3 & 77.6 & 79.5 \\
      SUAVE-Scratch        & 87.3 & 61.4 & 75.7 & 92.7 & 90.4 & 89.8 & 75.9 & 81.1 \\
      SUAVE-Cotrain        & \textbf{94.6} & \textbf{71.7} & 76.4 & 93.3 & \textbf{94.2} & \textbf{95.1} & \textbf{80.7} & \textbf{86.1} \\
      \bottomrule
  \end{tabular}
  \end{table}

\subsection{DOMINO for Dynamic Manipulation}
  \label{sec:domino}
\textbf{Setup.} DOMINO~\cite{fang2026towards} evaluates the manipulation of moving objects in simulation. We use its 25 single-arm Franka tasks and evaluate 100 episodes per task at each of the three dynamics levels: Level 1 where objects move at a constant velocity, Level 2 where objects move along polynomial trajectories with varying curvature, and Level 3 where objects move along piecewise segments with stochastic, discontinuous velocity. DOMINO reports three metrics: success rate (SR), a manipulation score (MS) that credits partial progress, and route completion (RC), which is MS before safety penalties. Every method is trained on all 18,231 Franka demonstrations for these tasks, covering clean and randomized scenes at all three levels, and the same checkpoint is evaluated at each level. All baselines are retrained by us using their authors' recipes. SUAVE-Cotrain is pretrained on DOMINO's non-Franka embodiments mixed with SSv2 and finetuned on the Franka demonstrations. Table~\ref{tab:domino_franka_levels} reports all three metrics with the standard error over 2,500 episodes per level.

\textbf{Discussion.} Table~\ref{tab:domino_franka_levels} answers the dynamic half of Q1. SUAVE-Cotrain has the highest success rate at every level, with 18.7\% at Level 1, 12.3\% at Level 2, and 8.2\% at Level 3. SUAVE-Scratch is second in success rate at every level. The strongest baseline is OpenVLA-OFT at 12.3\% on Level 1, followed by $\pi_0$-FAST at 10.8\%. The ordering differs drastically from LIBERO. $\pi_{0.5}$, the best model on LIBERO at 97.4\%, reaches 6.2\% on Level 1 and 1.7\% on Level 3, below both SUAVE variants. These methods degrade as the dynamics become less predictable. Cotrain improves on Scratch, showing that the effect of cross-embodiment pretraining has a small observed improvement at this sample size.

\begin{table*}[!t]
\caption{\textbf{DOMINO across dynamics levels (\%):} All rows are single checkpoints trained on the same Franka demonstrations and evaluated on the 25-task subset at 100 episodes per task. We report SR, MS, and RC as mean and
standard error over 2,500 episodes per level. Bold marks the best per column.}
\label{tab:domino_franka_levels}
\centering
\small
\setlength{\tabcolsep}{4pt}
\begin{tabular}{lccccccccc}
    \toprule
    & \multicolumn{3}{c}{Level 1} & \multicolumn{3}{c}{Level 2} & \multicolumn{3}{c}{Level 3} \\
    & \multicolumn{3}{c}{\footnotesize\textit{constant velocity}} & \multicolumn{3}{c}{\footnotesize\textit{polynomial}} & \multicolumn{3}{c}{\footnotesize\textit{stochastic, abrupt}} \\
    \cmidrule(lr){2-4} \cmidrule(lr){5-7} \cmidrule(lr){8-10}
    Method & SR & MS & RC & SR & MS & RC & SR & MS & RC \\
    \midrule
    UWM~\cite{zhu2025unified}              & \pmv{5.9}{0.5}  & \pmv{20.7}{0.4} & \pmv{28.5}{0.5} & \pmv{5.3}{0.5}  & \pmv{20.4}{0.4} & \pmv{29.0}{0.5} & \pmv{3.6}{0.4} & \pmv{17.3}{0.4} & \pmv{26.4}{0.4} \\
    $\pi_0$~\cite{black2024pi_0}           & \pmv{3.6}{0.4}  & \pmv{13.8}{0.4} & \pmv{21.6}{0.4} & \pmv{2.4}{0.3}  & \pmv{12.1}{0.3} & \pmv{20.0}{0.4} & \pmv{0.8}{0.2} & \pmv{10.4}{0.3} & \pmv{18.3}{0.4} \\
    $\pi_0$-FAST~\cite{pertsch2025fast}    & \pmv{10.8}{0.6} & \pmv{22.7}{0.5} & \pmv{30.3}{0.6} & \pmv{8.3}{0.6}  & \pmv{19.9}{0.5} & \pmv{27.9}{0.5} & \pmv{2.6}{0.3} & \pmv{15.1}{0.4} & \pmv{24.8}{0.4} \\
    $\pi_{0.5}$~\cite{intelligence2025pi_} & \pmv{6.2}{0.5}  & \pmv{16.1}{0.5} & \pmv{23.4}{0.5} & \pmv{4.2}{0.4}  & \pmv{13.9}{0.4} & \pmv{21.7}{0.4} & \pmv{1.7}{0.3} & \pmv{11.6}{0.3} & \pmv{19.8}{0.4} \\
    OpenVLA-OFT~\cite{kim2025fine}         & \pmv{12.3}{0.7} & \pmv{25.2}{0.5} & \pmv{32.8}{0.6} & \pmv{8.4}{0.6}  & \pmv{21.5}{0.5} & \pmv{29.4}{0.5} & \pmv{3.3}{0.4} & \pmv{15.5}{0.4} & \pmv{24.8}{0.4} \\
    PUMA~\cite{fang2026towards}            & \pmv{7.4}{0.5}  & \pmv{25.2}{0.4} & \pmv{33.3}{0.4} & \pmv{5.2}{0.4}  & \pmv{23.1}{0.4} & \pmv{31.4}{0.4} & \pmv{4.1}{0.4} & \pmv{\textbf{20.1}}{0.4} & \pmv{\textbf{29.4}}{0.4} \\
    \midrule
    SUAVE-Scratch (ours)                   & \pmv{17.0}{0.8} & \pmv{27.1}{0.6} & \pmv{33.7}{0.6} & \pmv{10.4}{0.6} & \pmv{21.4}{0.5} & \pmv{29.2}{0.6} & \pmv{7.6}{0.5} & \pmv{19.1}{0.5} & \pmv{27.5}{0.5} \\
    SUAVE-Cotrain (ours)                   & \pmv{\textbf{18.7}}{0.8} & \pmv{\textbf{29.0}}{0.7} & \pmv{\textbf{35.4}}{0.7} & \pmv{\textbf{12.3}}{0.7} & \pmv{\textbf{24.1}}{0.6} & \pmv{\textbf{32.4}}{0.6} & \pmv{\textbf{8.2}}{0.6} &
\pmv{20.0}{0.5} & \pmv{28.7}{0.6} \\
    \bottomrule
\end{tabular}
\end{table*}

\section{Conclusion}
We presented SUAVE, a masked diffusion transformer that represents language, video, and actions as discrete tokens in one sequence. Choosing which tokens to mask at inference turns the same weights into a world model, a robot policy, or a video-action model with no change to the architecture or training objective. On a real robot, the joint model generates two visual subgoals and an action chunk at a rate sufficient for closed-loop control. Our model is competitive with specialized policies on both static and dynamic manipulation benchmarks, and the same weights predict long-horizon video comparably to dedicated video models. A controlled study on a real robot shows that pretraining on robot video and co-training on action-free human video more than doubles task scores and preserves that gain under distribution shift.

\textbf{Limitations.} At our scale, co-training yields higher task scores than pretraining alone, however the difference is not statistically separable. More diverse action-free corpora would provide a clearer signal. We use a frozen MAGViT-v2 tokenizer, which bounds the reconstruction quality of predicted frames. Finetuning it directly on robot data could improve visual quality. Our token sequences have a fixed length. Variable length decoding would allow the model to adapt its horizon to the task. We evaluate a single 7-DoF arm with an end-effector action space and do not test other embodiments. Finally, although our backbone is a language model, we do not generate language and do not evaluate semantic generalization. We leave these directions for future work.

\bibliographystyle{IEEEtran}
\bibliography{references}

\newpage

\vfill

\end{document}